\documentclass[letterpaper]{article} 
\usepackage[draft]{aaai2026}  
\usepackage{times}  
\usepackage{helvet}  
\usepackage{courier}  
\usepackage[hyphens]{url}  
\usepackage{graphicx} 
\usepackage{natbib}  
\usepackage{caption} 
\usepackage{algorithm}
\usepackage{algorithmic}
\usepackage{bm}

\usepackage{amssymb}

\usepackage{booktabs} %
\usepackage{tikz}
\usepackage{pgfplots}
\usetikzlibrary{positioning}
\usetikzlibrary{calc}
\usepackage{graphicx}
\usepackage{subcaption}
\usepackage{etoolbox,lipsum}
\usepackage{xcolor}
\usepackage{afterpage}
\definecolor{lightred}{HTML}{F08080}    
\definecolor{lightorange}{HTML}{FDBA74} 
\definecolor{lightyellow}{HTML}{FDE68A} 
\definecolor{skyblue}{rgb}{0.53, 0.81, 0.98}
\usepackage{colortbl}
\usepackage{etoolbox,lipsum}

\usepackage[utf8]{inputenc}
\usepackage{amsmath}
\usepackage{tikz}
\usetikzlibrary{positioning}
\usetikzlibrary{calc}
\usepackage{caption}

\usepackage{cleveref}
\crefname{figure}{Fig.}{Figs.}
\crefname{table}{Tab.}{Tabs.}
\crefname{section}{Sec.}{Secs.}
\crefname{equation}{Eq.}{Eqs.}
\usepackage{newfloat}
\usepackage{listings}
\DeclareCaptionStyle{ruled}{labelfont=normalfont,labelsep=colon,strut=off} 
\floatstyle{ruled}
\newfloat{listing}{tb}{lst}{}
\floatname{listing}{Listing}
\newcommand{\br}{\mathbf{r}}

\usepackage{pifont}

\newcommand{\cmark}{\ding{51}}
\newcommand{\xmark}{\ding{55}}

\makeatletter
\DeclareRobustCommand\onedot{\futurelet\@let@token\@onedot}
\def\@onedot{\ifx\@let@token.\else.\null\fi\xspace}

\makeatother

\title{From Restoration to Reconstruction: Rethinking 3D Gaussian Splatting for Underwater Scenes}
\author{
    Guoxi Huang\textsuperscript{\rm 1}\equalcontrib,
    Haoran Wang\textsuperscript{\rm 1}\equalcontrib,
    Zipeng Qi\textsuperscript{\rm 2},
    Wenjun Lu \textsuperscript{\rm 3}, \\
    David~Bull\textsuperscript{\rm 1},
    Nantheera Anantrasirichai\textsuperscript{\rm 1},
}
\affiliations{
    \textsuperscript{\rm 1} Visual Information Laboratory, 
 University of Bristol,  Bristol, UK\\
     \textsuperscript{\rm 2} Beijing University of Aeronautics and Astronautics, Beijing, China\\
     \textsuperscript{\rm 3} The University of Sydney

}

\usepackage{bibentry}

\begin{document}

\maketitle

\begin{abstract}
Underwater image degradation poses significant challenges for 3D reconstruction, where simplified physical models often fail in complex scenes. We propose \textbf{R-Splatting}, a unified framework that bridges underwater image restoration (UIR) with 3D Gaussian Splatting (3DGS) to improve both rendering quality and geometric fidelity. 
Our method integrates multiple enhanced views produced by diverse UIR models into a single reconstruction pipeline. During inference, a lightweight illumination generator samples latent codes to support diverse yet coherent renderings, while a contrastive loss ensures disentangled and stable illumination representations.
Furthermore, we propose \textit{Uncertainty-Aware Opacity Optimization (UAOO)}, which models opacity as a stochastic function to regularize training. This suppresses abrupt gradient responses triggered by illumination variation and mitigates overfitting to noisy or view-specific artifacts. Experiments on Seathru-NeRF and our new BlueCoral3D dataset demonstrate that R-Splatting outperforms strong baselines in both rendering quality and geometric accuracy.

\end{abstract}

\section{Introduction}

Ocean exploration is gaining momentum due to its applications in underwater archaeology, geology, and marine sciences. However, it remains hindered by technical limitations, high costs, and a shortage of skilled divers. As a result, 3D scene reconstruction has become vital, enabling efficient, scalable, and remote analysis of underwater environments.

Yet, achieving accurate 3D reconstruction in underwater environments remains highly challenging due to the unique optical properties of water. Unlike terrestrial scenes, underwater imagery suffers from depth-dependent color attenuation, scattering-induced haze, and non-uniform illumination, all of which impair geometric and photometric consistency across views.

Reconstruction methods such as NeRF~\cite{mildenhall2021nerf,muller2022instant} and 3D Gaussian Splatting~\citep{kerbl20233d,yu2024mip} typically assume clear-air conditions, and thus lack the modeling capacity to handle the complex light transport in participating media. When directly applied to underwater images, these methods often fail to recover fine structures or maintain consistent appearance, resulting in significant reconstruction artifacts.

\begin{figure}
    \centering
    \includegraphics[width=1.\linewidth]{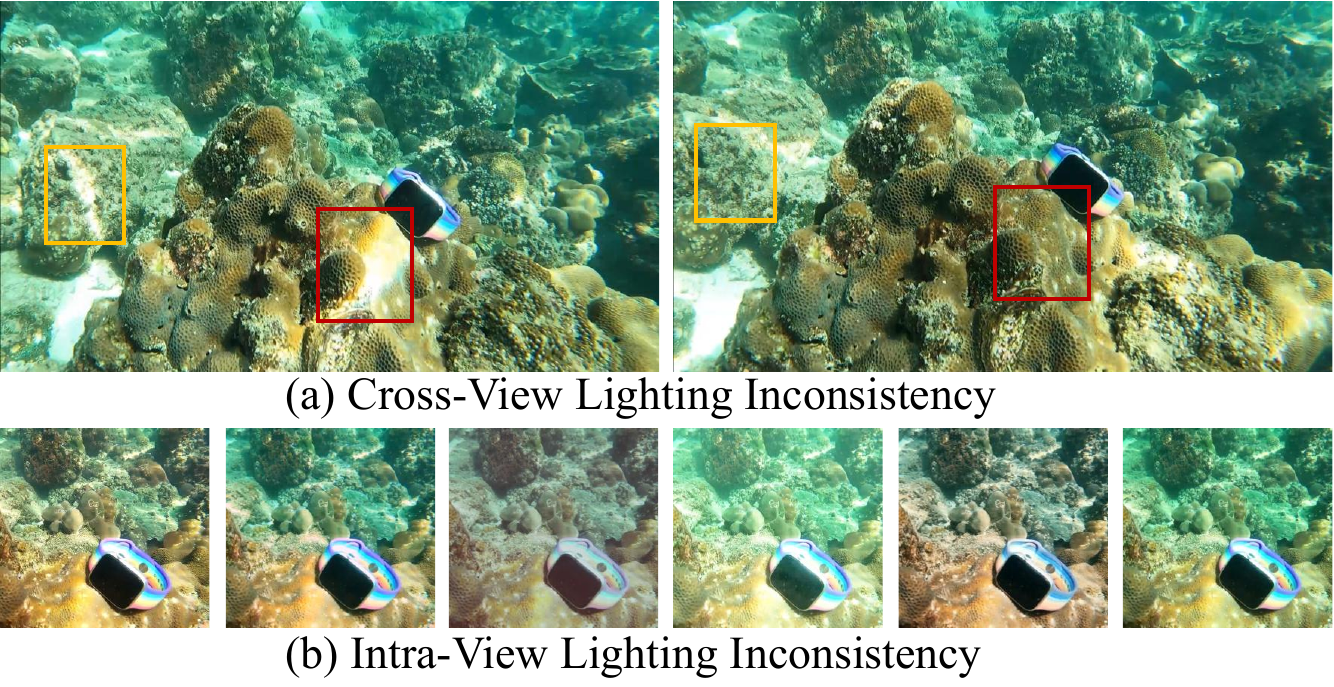}
    \caption{Illustration of illumination inconsistencies in underwater scenes.
(a) Cross-view: the same object appears with different shading across viewpoints due to viewpoint-dependent lighting and refraction.(b) Intra-view: different restoration models applied to the same view introduce inconsistent illumination effects.}
    \label{fig:reflect}
\end{figure}

Recent methods~\cite{levy2023seathru,li2025watersplatting,wang2025uw,jiang2025rusplatting} incorporate simplified physical models into NeRF and 3DGS frameworks to account for underwater light attenuation and backscattering, thereby improving rendering quality to some extent. However, these approaches are fundamentally limited by the oversimplified assumptions in their physical models—for instance, the common neglect of forward scattering~\cite{Tan:Visibility:2008} and light reflection leads to failure in complex underwater scenes.

Underwater image restoration (UIR) techniques have demonstrated strong capabilities in enhancing degraded underwater images. However, their integration into 3D reconstruction pipelines remains largely unexplored. Existing neural rendering methods typically operate directly on raw underwater inputs, often ignoring degradations such as color distortion, haze, or contrast loss. As a result, reconstructions may suffer from inaccurate geometry and poor appearance fidelity.

No single UIR model is universally optimal—different methods target specific degradations, such as color correction, haze removal, or contrast enhancement. Moreover, due to the ill-posed nature of underwater restoration, deep learning-based methods—particularly generative models—often produce diverse outputs from the same input, resulting in \textit{intra-view lighting inconsistencies} (see~\cref{fig:reflect}, bottom). Such variations can negatively impact the performance of downstream 3D reconstruction.

Training a separate 3D model for each enhanced version is not only computationally expensive but also incurs significant storage overhead. This motivates a unified framework that can jointly handle diverse restorations while maintaining geometry consistency. By leveraging complementary strengths—e.g., better visibility, sharper textures, or corrected color balance—we can enrich the reconstruction process with varied enhancement perspectives that reveal more reliable geometry and appearance cues.

Meanwhile, underwater scenes often exhibit cross-view illumination inconsistencies due to view-dependent light reflectance. As shown in~\cref{fig:reflect} (Top), the same object may appear differently across views depending on its orientation to the water surface and lighting. This makes it difficult for 3D models to learn consistent geometry and appearance.

To address these challenges, we propose \textit{Restoration Splatting} (R-Splatting), a unified 3D reconstruction framework that integrates 3D-level uncertainty modeling and a neural field to condition illumination within the 3D Gaussian Splatting (3DGS) paradigm.

To resolve cross-view lighting inconsistencies, we introduce \textit{Uncertainty-Aware Opacity Optimization} (UAOO;~\cref{sec:Uncertainty-Aware Opacity Optimization}), which uses a random field to model per-point uncertainty. This enables our method to suppress artifacts while preserving fine-grained geometry.

To address intra-view lighting inconsistencies caused by different UIR methods, R-Splatting also incorporates a neural field that learns a conditional representation of image illumination, allowing the 3D model to adapt to diverse enhancement styles. This design supports rendering under multiple plausible lighting conditions. To further disentangle lighting cues, we employ a contrastive learning objective (\cref{sec:view-invariant}) that separates appearance embeddings across styles.

By modeling both visual diversity and geometric consistency, R-Splatting bridges UIR and neural 3D reconstruction, enabling scalable, coherent scene modeling in challenging underwater environments.

\noindent \paragraph{Our contributions can be summarized as follows:}
\begin{itemize}
    \item We propose \textbf{R-Splatting}, a unified framework that fuses multiple UIR results into a single 3DGS model, enhancing reconstruction quality through diverse visual cues.

    \item We introduce \textbf{UAOO} to model per-point uncertainty via stochastic opacity, improving robustness to illumination inconsistencies.

    \item We release \textbf{BlueCoral3D}, a dataset with dynamic lighting, and validate our method’s effectiveness through extensive experiments.

    \item Our approach is \textbf{flexible}, supporting future UIR advances without requiring model changes.
\end{itemize}

\section{Related Works}
\noindent \paragraph{Underwater Image Restoration.}
Underwater image restoration (UIR) methods aim to address common issues in underwater photography, such as blurring effects, floating particles, and color cast, while also improving the overall visual quality by enhancing contrast and brightness. Existing UIR approaches can be broadly categorized into physical model-based methods~\cite{he2010single,Yang:Low:2011,Chiang:Underwater:2012,Wen:Single:2013,Galdran:Automatic:2015,peng2018generalization,Liang:GUDCP:2022,zhang2025underwater} and deep learning-based methods~\cite{tang2023underwater,Li:Underwater:2021,peng2023u,Guan:WaterMamba:2024,huang2025bayesian,huang2025bvi,huang2025visual,lin2024pixmamba,peng2025adaptive,malyugina2025marine}. Recent studies demonstrate that deep learning-based methods outperform traditional physical model-based techniques in both restoration quality and robustness. Hence, in this work, we adopt off-the-shelf deep learning UIR models to enhance underwater images before further processing.

\begin{figure*}[h]
    \centering
    \includegraphics[width=1.\linewidth]{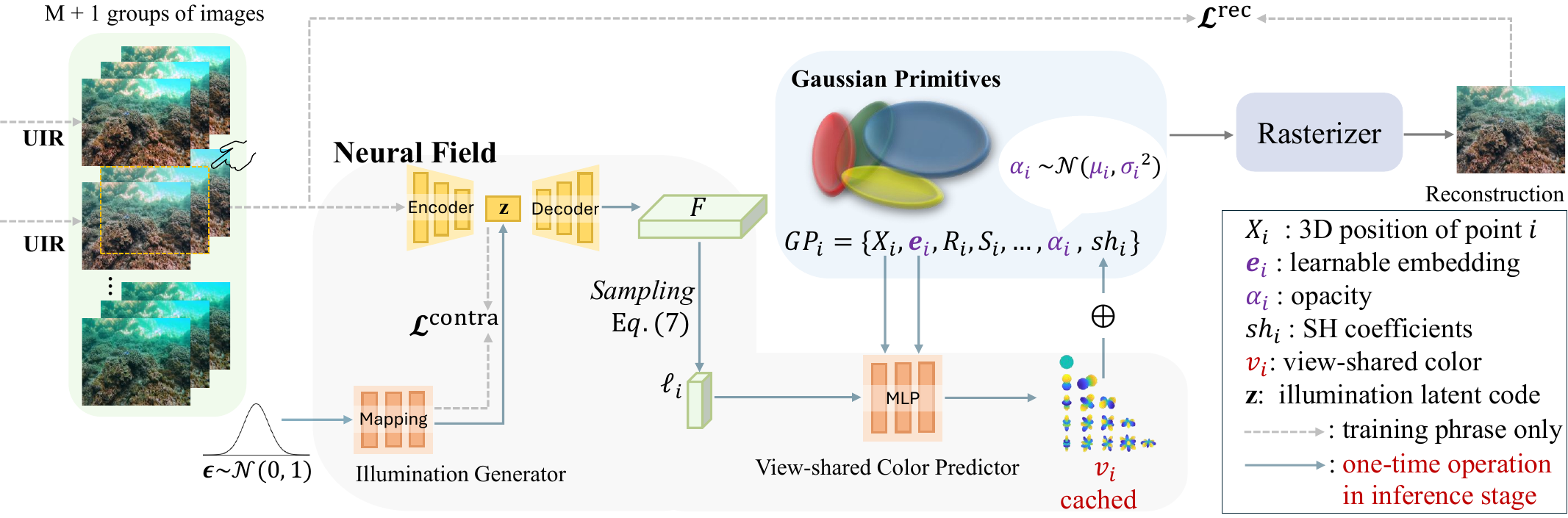}
\caption{
\textbf{Overview of the proposed \textbf{R-Splatting} pipeline.} 
(1) We first apply multiple UIR models to obtain $M{+}1$ groups of enhanced images. 
(2) A neural field encodes each image into a latent code $\mathbf{z}$, regularized by a contrastive loss to remove view-dependent cues. 
An illumination generator predicts $\mathbf{z}$ at inference. 
The latent is bilinearly sampled into point-wise features and fused with Gaussian attributes to produce a view-shared color. 
(3) An uncertainty-aware renderer models opacity as a distribution to suppress artifacts and improve multi-view consistency.
}
    \label{fig:framework}
\end{figure*}

\begin{figure}[h]
    \centering
    \includegraphics[width=1.\linewidth]{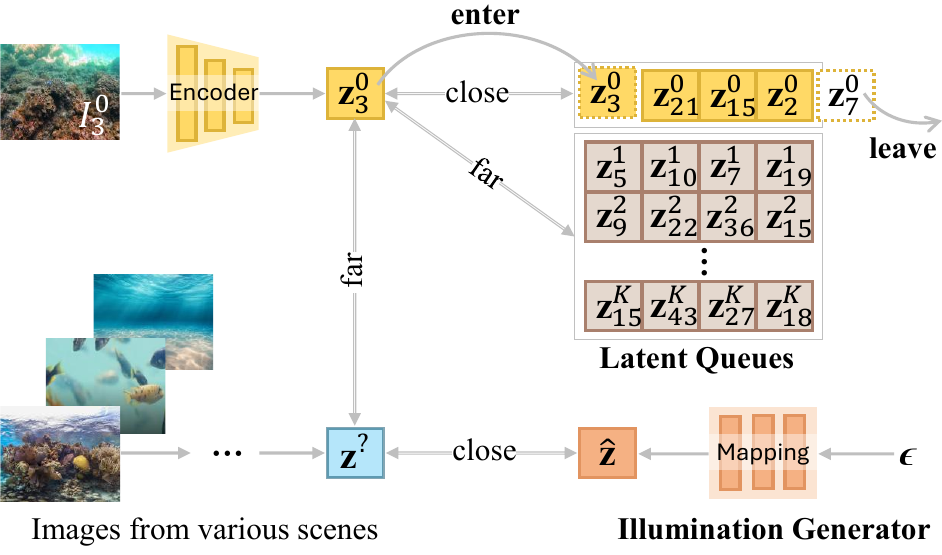}
    \caption{ \textbf{Learning view-invariant latent code with dynamic latent queues.} For each restoration style, the newly generated latent code is added to the front of its corresponding subqueue, while the oldest code at the end is removed to maintain a fixed queue length. We maximize the similarity between latent codes of different views under the same restoration style, and minimize the similarity between those from different restoration styles.}
    \label{fig:contra_loss}
\end{figure}

\noindent \paragraph{Underwater 3D Reconstruction.} ScatterNeRF~\citep{ramazzina2023scatternerf} addressed participating media by separating attenuation from geometry, which applies to underwater settings. SeaThru-NeRF~\citep{levy2023seathru} introduced per-ray parameters based on a revised underwater formation model, later extended by SP-SeaNeRF~\citep{CHEN2024104025} with learnable illumination. WaterNeRF~\citep{Sethuraman:WaterNeRF:2023} incorporated light transport and color correction via optimal transport, while WaterHE-NeRF~\citep{ZHOU2025102770} applied a Retinex-inspired field for color compensation. For dynamic scenes, UWNeRF~\citep{10656460} separates motion via masks, and AquaNeRF~\citep{gough2025aquanerf} improves transmittance modeling per ray.

More recently, 3DGS has been adapted for underwater scenarios. Z-Splat~\citep{qu2024z} fuses sonar and RGB to address sparse-view issues but lacks medium modeling. RecGS~\citep{zhang2024recgs} applies filtering and recurrence to improve view quality without modeling scattering. Gaussian Splashing~\citep{mualem2024gaussian} embeds scattering directly in CUDA for efficiency.
Aquatic-GS~\citep{liu2024aquatic} and SeaSplat~\citep{yang2024seasplat} extend 3DGS by incorporating haze-aware image formation models. Building on this direction, UW-GS~\citep{wang2025uw} and WaterSplatting~\cite{li2025watersplatting} combine physical priors with MLPs to estimate underwater attenuation coefficients for rendering restoration. SWAGSplatting~\cite{jiang2025swagsplatting} incorporates high-level semantics into the reconstruction process by leveraging Grounded-SAM~\cite{ren2024grounded}.

However, no prior work has attempted to unify underwater image restoration and 3D Gaussian Splatting, a gap we address in this paper.

\subsection{Preliminaries: 3DGS}
3DGS~\cite{kerbl20233d} represents a 3D scene as a set of Gaussian primitives ${\mathcal{G}_i}$. Each Gaussian $\mathcal{G}_i$ is defined by a mean position $X_i$, a covariance matrix $\Sigma_i$, an opacity $\alpha_i$, and a view-dependent color $c_i$ encoded via spherical harmonics (SH). The covariance matrix $\Sigma_i$ is decomposed into a rotation matrix $R_i \in \mathbb{R}^{3 \times 3}$ and a scaling matrix $S_i \in \mathbb{R}^{3 \times 3}$ to enable differentiable optimization:
\begin{equation}
\Sigma_i = R_i S_i S_i^{T} R_i^{T}.
\end{equation}

During rendering, each 3D Gaussian is projected onto the image plane~\cite{zwicker2001surface} using a viewing transformation $W$. 
This projection yields a corresponding 2D Gaussian, whose image-space covariance matrix $\Sigma'_i$ is given by:
\begin{equation}
\Sigma'_i = J W \Sigma_i W^{T} J^{T} \enspace ,
\end{equation}
where $J$ is the Jacobian matrix of an affine approximation to the projective transformation.

To render an image from a given camera pose, we compute the color of each pixel via alpha compositing: 
For each pixel, the Gaussians are sorted in front-to-back order based on their distances to the image plane.
The 2D Gaussian’s mean $X'_i$  is obtained by projecting $X_i$ into the camera coordinate system using $W$.
Each Gaussian contributes a view-dependent color $c_i$. The final pixel color $C$ is computed as:
\begin{align}
\alpha'_i &= \alpha_i \exp\left( -\tfrac{1}{2} (x - X'_i)^T (\Sigma'_i)^{-1} (x - X'_i) \right), \\
C &= \sum_{i} c_i \alpha_i \prod_{j=1}^{i-1} (1 - \alpha_j),
\end{align}
where $\alpha_i$ signify the opacity of point $i$, which is a learnable parameter, and $\alpha_i$ represents the blending weight of the $i$-th Gaussian.
To model view-dependent effects such as specularities, each Gaussian stores a base set of spherical harmonics (SH) coefficients, denoted as $\text{sh}_i$, which are used to compute the view-dependent color $c_i$ based on the viewing direction.

\section{Method}
Given $N$ multi-view underwater images, we apply multiple UIR models to produce $M$ enhanced versions, resulting in $M+1$ image sets per view (including the original). Our goal is to reconstruct all $M+1$ sets, denoted as $\{\{ \mathbf{I}^m_n \}_{n=1}^{N}\}_{m=0}^{M}$, into a unified GS model.
However, variations in illumination and geometry across these sets—especially those produced by generative UIR models—can introduce multi-view inconsistencies, leading to blurred textures and inaccurate lighting representations.
To overcome this, we propose Restoration Splatting (R-Splatting). The overall pipeline is shown in~\cref{fig:framework}.

\subsection{View-shared Color Modeling}
\label{sec:Conditional GS}
To address \textit{intra-view lighting inconsistencies} introduced by different UIR models, we integrate a neural field into 3DGS to model view-shared color. 
Intuitively, by assigning a shared illumination to all Gaussian primitives, we can achieve illumination-consistent rendering across views. 

We first employ a tiny autoencoder~\cite{taesd2023} to extract a latent code $\mathbf{z}$ from image $I$, and then further map it onto feature maps $F$: 
\begin{equation}
\begin{aligned}
    \mathbf{z}&= \mathcal{E}(I), \\
    F &= \mathcal{D}(\mathbf{z}).
\end{aligned}
\end{equation}
Then, given the known camera pose, each 3D point is projected to the 2D image space for feature sampling, as illustrated below:
\begin{equation}
\begin{aligned}
    \bm \ell_i = \mathrm{BilinearSample} \left( F, \frac{1}{z_i}
\begin{bmatrix}
u_i \\
v_i
\end{bmatrix} \right), \\
\quad \text{where} \quad
\begin{bmatrix}
u_i \\
v_i \\
z_i
\end{bmatrix}
= P \cdot
\begin{bmatrix}
X_i \\
1
\end{bmatrix}.
\end{aligned}
\end{equation}
Each 3D point $X_i$ is first converted into homogeneous coordinates and projected onto the 2D image plane using the projection matrix $P = K W$, where $K$ denotes the camera intrinsics. This yields a 2D coordinate $\left( \frac{u_i}{z_i}, \frac{v_i}{z_i} \right)$. We then sample the corresponding feature $\ell_i$ from the feature map $F$ at the pixel location using bilinear interpolation. Such that, each Gaussian point has its own illumination feature vector.

Subsequently, the point-wise latent code $\bm{\ell}_i$ is fed into a lightweight MLP, along with a learnable embedding $\mathbf{e}_i$ for each Gaussian $i$, the 3D position $X_i$ to predict the view shared color encoded as a set of SH coefficients:
\begin{equation}
\bm v_i = \mathrm{MLP}(\bm{\ell}_i, \mathbf{e}_i, X_i)
\end{equation}
The learnable embedding $\mathbf{e}_i$ plays a role analogous to parametric encodings~\cite{muller2022instant}, improving the expressiveness and capacity of a lightweight MLP without significantly increasing computational cost.
The final color are then computed as:
\begin{equation}
    c_i(\br) = \mathrm{SH}(\br; sh_i + \bm v_i),
\end{equation}
where $\mathrm{SH}(\cdot)$ denotes the standard SH basis evaluation with the given coefficients.

Note that during multi-view evaluation, the entire \textbf{Neural Field} is executed only once to produce the view-shared color representation. Subsequently, multiple views can be rendered by simply updating the view direction (\textit{i.e.,}  $R_i$ and $S_i$) in the rasterizer, introducing only marginal overhead to the overall rendering latency. 

\begin{figure*}[h]
    \centering
    \includegraphics[width=1.\linewidth]{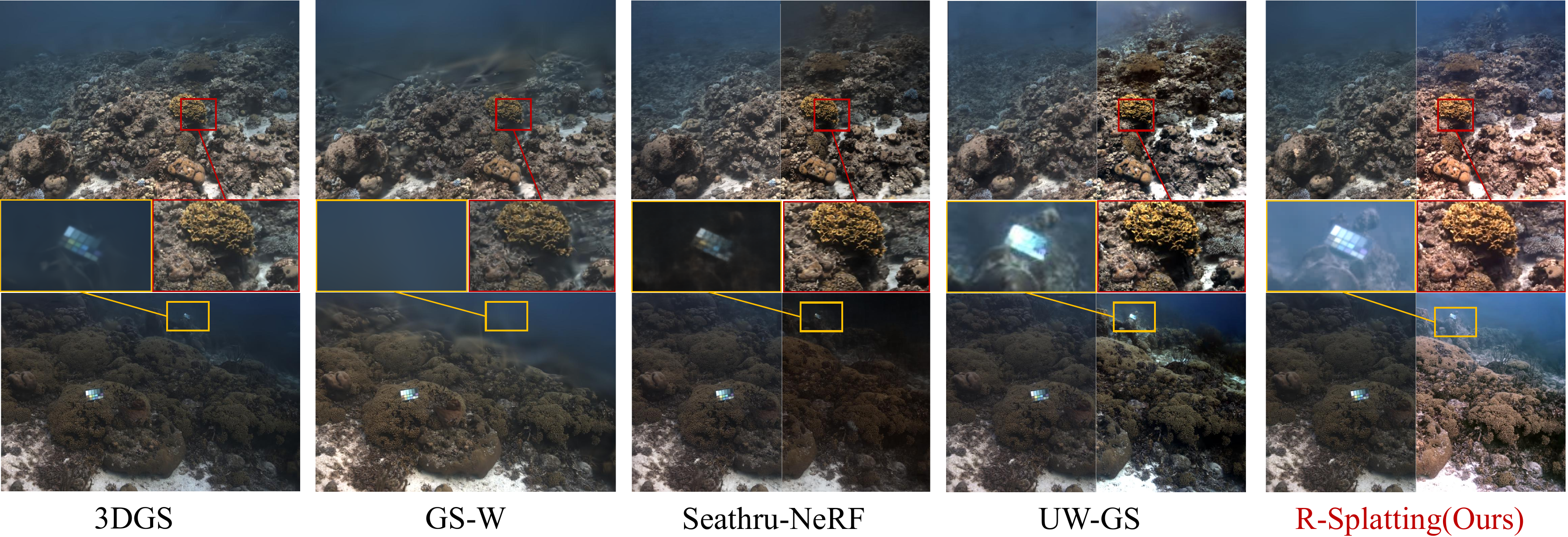}
    \caption{\textbf{Comparison on the SeaThru-NeRF dataset.} Right halves show the restored outputs from models with restoration capabilities (columns 3–5). Our method shows higher visibility than the baseline methods in the renderings.}
    \label{fig:visual_seathru_nerf}
\end{figure*}

Concurrent works such as~\cite{kulhanek2024wildgaussians,wang2025uw} also use a neural field for color prediction, but its view-dependent design requires per-view inference, leading to higher rendering latency.

\paragraph{Learning View-invariant Latent.}
\label{sec:view-invariant}

When rendering multiple views from different angles, our conditional GS model uses a single latent code. This requires that the latent code be free of view-dependent information; otherwise, inconsistencies in lighting and appearance may arise across different viewpoints.
However, without explicit constraints, the encoder tends to inadvertently encode view-dependent information into the latent code.

To address this issue, we introduce a contrastive loss~\cite{he2020momentum}, as depicted in~\cref{fig:contra_loss}. 
The goal is to pull latent codes from images with similar underwater ``styles" together while pushing apart those from different styles, as follows:
\begin{equation}
\mathcal{L}^\mathrm{contra} = -\log 
\frac{
    \sum\limits_{n} \exp(\mathbf{z}_{j}^m \cdot \mathbf{z}_n^m / \tau)
}{
    \sum\limits_{m} \sum\limits_{n} \exp(\mathbf{z}_{j}^m \cdot \mathbf{z}_n^m / \tau) + \exp(\mathbf{z}_{j}^m \cdot \tilde{\mathbf{z}} / \tau) 
},
\label{eq:contra loss}
\end{equation}
where $\tau$ is a temperature hyperparameter set to 0.07. $\tilde{\mathbf{z}}$ denotes a latent code sampled from a random clean scene, which helps the encoder generalize to illumination conditions beyond those present in the reconstructed scene.

To provide rich sets of positive and negative examples, we maintain dynamic queues of recent latent codes for different restoration styles.

\noindent \paragraph{Illumination Generator.}
To enable diverse appearance synthesis without relying on image-based latent extraction at inference time, we introduce a lightweight illumination generator that learns to sample plausible lighting embeddings directly from noise. This module comprises five convolutional layers with ReLU activations, mapping a random noise vector to an illumination latent in the same embedding space as the autoencoder's output. Both the input and output have the shape of latent code $\mathbf{z}$. 
The generator is supervised by maximizing the similarity between its output and the autoencoder-generated latent codes. This loss is integrated into the contrastive objective described in~\cref{eq:contra loss}.

\begin{table*}
    \centering
    \small
    \setlength{\tabcolsep}{4pt}
    \begin{tabular}{l|c|ccc|ccc|ccc|ccc}
         & \textbf{FPS /  GPU hrs.} & \multicolumn{3}{c|}{\textbf{Curaçao}} & \multicolumn{3}{c|}{\textbf{Panama}} & \multicolumn{3}{c|}{\textbf{IUI3}} & \multicolumn{3}{c}{\textbf{Japanese Gardens}} \\
         &  & PSNR & SSIM & LPIPS & PSNR & SSIM & LPIPS & PSNR & SSIM & LPIPS & PSNR & SSIM & LPIPS \\
        \hline
        SeaThru-NeRF   & $<$ 1 / 11.8 & \cellcolor{lightyellow}30.00 & 0.870 & 0.215 & \cellcolor{lightyellow}27.82 & \cellcolor{lightyellow}0.834 & \cellcolor{lightyellow}0.226 & 25.92 & 0.787 & 0.294 & 21.73 & 0.768 & 0.246 \\
        3DGS  & 163 / 0.38 & 29.28 & \cellcolor{lightyellow}0.923& \cellcolor{lightyellow}0.202 & 23.68& 0.749 & 0.250 &
        \cellcolor{lightorange}29.76 & \cellcolor{lightyellow}0.875 & \cellcolor{lightorange}0.121 & 21.45 & \cellcolor{lightyellow}0.852 & \cellcolor{lightorange}0.206 \\
        WildGaussian   & 73 / 1.30 & 29.52 & 0.880 & 0.313 & 24.94 & 0.756 & 0.395 & 28.34 & 0.870 & 0.177 & \cellcolor{lightyellow}22.08 & 0.839 & 0.312 \\
        GS-W           & 42 / 1.69 & 24.20 & 0.843 & 0.382 & 24.30 & 0.748 & 0.387 & 27.64 & 0.863 & 0.263 & 21.46 & 0.850 & 0.244 \\
        UW-GS          & 30 / 0.38 & \cellcolor{lightorange}31.77 & \cellcolor{lightorange}0.943 & \cellcolor{lightred}0.144 & \cellcolor{lightorange}31.79 & \cellcolor{lightred}0.936 & \cellcolor{lightorange}0.116 & \cellcolor{lightyellow}28.65 & \cellcolor{lightorange}0.933 & \cellcolor{lightyellow}0.125 & \cellcolor{lightorange}23.05 & \cellcolor{lightorange}0.860 & \cellcolor{lightred}0.190 \\
        \textbf{R-Splatting (Ours)}       & 107 / 0.86 & \cellcolor{lightred}32.98 & \cellcolor{lightred}0.956 & \cellcolor{lightorange}0.163 & \cellcolor{lightred}32.52 & \cellcolor{lightyellow}0.930 & \cellcolor{lightred}0.107 & \cellcolor{lightred}30.15 & \cellcolor{lightred}0.947 & \cellcolor{lightred}0.105 & \cellcolor{lightred}24.03 & \cellcolor{lightred}0.868 & \cellcolor{lightyellow}0.211
    \end{tabular}%
    \caption{\textbf{Quantitative evaluation on the SeaThru-NeRF dataset} in terms of rendering speed (FPS)$\uparrow$, training time (GPU hrs)$\downarrow$, PSNR$\uparrow$, SSIM$\uparrow$, and LPIPS$\downarrow$. The \colorbox{lightred}{best}, \colorbox{lightorange}{second-best}, and \colorbox{lightyellow}{third-best} results are highlighted. All FPS and runtime values are measured on an NVIDIA RTX 4090.}
    \label{tab:seathru_compare}
\end{table*}

\begin{table}
    \centering
    \small
    \setlength{\tabcolsep}{3pt}
    \begin{tabular}{l|ccc|ccc}
         & \multicolumn{3}{c|}{\textbf{Scene 1}} & \multicolumn{3}{c}{\textbf{Scene 2}} \\
        & PSNR & SSIM & LPIPS & PSNR & SSIM & LPIPS \\
        \hline
        Seathru-NeRF & 15.92 & 0.437 & 0.604 & 20.66& 0.619 & 0.541 \\
        3DGS              & 21.17 & \cellcolor{lightorange}0.772 & \cellcolor{lightred}0.285 & 24.45 & 0.848 & \cellcolor{lightyellow}0.271 \\
        UW-GS             & \cellcolor{lightyellow}21.33 & \cellcolor{lightyellow}0.752 & \cellcolor{lightorange}0.290 & \cellcolor{lightyellow}24.96 & \cellcolor{lightorange}0.851 & \cellcolor{lightorange}0.265 \\
        WaterSplatting    & 20.80 & 0.721 & 0.348 & 24.45 & 0.832 & 0.299 \\
        WildGaussian      & \cellcolor{lightorange}21.76 & 0.748 & 0.387 & \cellcolor{lightorange}24.98 & \cellcolor{lightyellow}0.849 & 0.304 \\
        \textbf{R-Splatting (our)} & \cellcolor{lightred}23.04 & \cellcolor{lightred}0.780 & \cellcolor{lightyellow}0.301 & \cellcolor{lightred}26.31 & \cellcolor{lightred}0.860 & \cellcolor{lightred}0.253 \\
    \end{tabular}%
    \caption{\textbf{Quantitative evaluation on the new BlueCoral3D dataset} in terms of PSNR$\uparrow$, SSIM$\uparrow$, and LPIPS$\downarrow$.}
    \label{tab:blue3d_compare}
\end{table}

\subsection{Uncertainty-Aware Opacity Optimization (UAOO)}
\label{sec:Uncertainty-Aware Opacity Optimization}

In scenes with inconsistent illumination across views, artifacts often emerge as Gaussian primitives overfit to view-specific appearance variations, causing excessive opacity via large gradients on $\alpha$. Prior works~\cite{kerbl20233d,kulhanek2024wildgaussians,zhang2024gaussian} mitigate this by periodically resetting opacities or applying 2D mask-based filtering. However, these approaches risk removing real structures or fail to capture cross-view geometric inconsistencies.

To address \textit{cross-view lighting inconsistencies}, we explicitly model uncertainty at the 3D point level by introducing a stochastic opacity perturbation during training. This strategy regularizes per-Gaussian opacity learning and suppresses abrupt gradient responses induced by dynamic illumination.

Specifically, instead of computing opacity from a fixed deterministic parameter, we model it as a stochastic function of a Gaussian-distributed latent variable. The compositing weight is computed as a perturbed sigmoid:
\begin{equation}
    \alpha_i^{\text{train}} = \mathcal{S}(\mu_i + \sigma_i \cdot \epsilon), \quad \text{with} \quad \epsilon \sim \mathcal{N}(0, 1), 
    \label{eq:sto opacity}
\end{equation}
where $\mathcal{S}$ denotes the sigmoid function, and $\mu_i$ and $\sigma_i$ represent the mean and standard deviation of the learned opacity distribution.
This means that we no longer learn $\alpha_i$ directly; instead, we model it implicitly through two separate pathways, $(\mu_i, \sigma_i)$, thereby:
\begin{itemize}
    \item Decomposing the gradient flow response of $\alpha_i$ into a mean pathway ($\mu_i$) and a perturbation pathway ($\sigma_i$):
    \begin{equation}
\frac{\partial \mathcal{L}}{\partial \mu_i} \propto \frac{\partial \mathcal{L}}{\partial \alpha_i} \cdot \mathcal{S}'(\mu_i + \sigma_i \cdot \epsilon),
\quad
\frac{\partial \mathcal{L}}{\partial \sigma_i} \propto \frac{\partial \mathcal{L}}{\partial \alpha_i} \cdot \mathcal{S}'(\mu_i + \sigma_i \cdot \epsilon) \cdot \epsilon.
\end{equation}
\item Enabling the model to not only represent the opacity value itself, but also to capture its sensitivity to illumination-induced uncertainty.
\end{itemize}

This stochastic formulation introduces variability into the gradient path by modulating the same residual $\frac{\partial \mathcal{L}}{\partial \alpha_i}$ with different samples of the noise term $\epsilon$. As a result, the model is prevented from consistently amplifying updates along any single direction based on a single noisy sample. This effectively suppresses overreaction to large residuals arising from outlier views or lighting inconsistencies, encouraging more stable opacity learning across varying illumination. Conceptually, our idea resembles gradient noise regularization~\cite{neelakantan2015adding}.

During inference, we replace the stochastic opacity with its expected value via a closed-form approximation:
\begin{equation}
\begin{aligned}
 \alpha_i^{\text{test}} &= \mathbb{E}_{\epsilon} \left[ \sigma(\mu_i + \sigma_i \cdot \epsilon) \right], \\
 \alpha_i^{\text{test}} &\approx \sigma\left( \frac{\mu_i}{\sqrt{1 + \frac{\pi^2}{8} \sigma_i^2}} \right).
 \end{aligned}
\end{equation}
This yields a smooth, deterministic opacity that accounts for uncertainty. Intuitively, points with higher uncertainty (\textit{i.e.,}  larger $\sigma_i$) receive softer opacity values, reducing the influence of ambiguous or low-confidence regions in the final rendering. Conversely, confident points (\textit{i.e.,}  small $\sigma_i$) maintain sharper transitions, ensuring structure preservation. 

To encourage the model to reflect uncertainty in ambiguous regions, we use a regularization loss: 
$
    \mathcal{L}^\text{ucn} =  - \sum_i |\sigma_i|. 
$

\begin{figure*}[h]
    \centering
    \includegraphics[width=1.\linewidth]{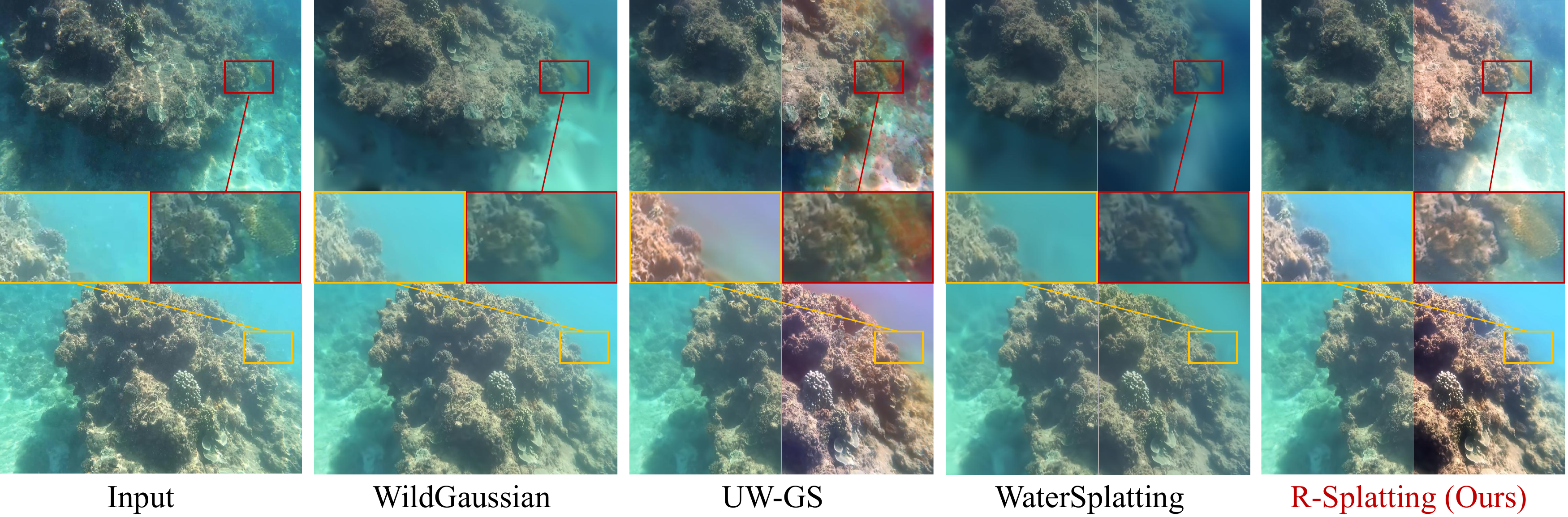}
    \caption{\textbf{Comparison on the BlueCoral3D dataset.}
    Restored renderings (right halves) show that UW-GS and WaterSplatting fail to recover accurate colors, due to unmodeled light reflections in shallow water.
    }
    \label{fig:visual_bluecoral3d}
\end{figure*}

\section{Experiments}

\noindent \textbf{Datasets.}
Due to the challenges of illumination inconsistency, prior works have typically focused on small-scale datasets with around 20 front-facing images per scene. In contrast, our method robustly handles complex lighting variations, enabling large-scale reconstruction. To validate this, we introduce \textbf{BlueCoral3D}, a new dataset comprising two underwater scenes with dynamic illumination. Each scene contains approximately 500 high-resolution (1080p) images captured through a full 360-degree sweep, naturally introducing diverse lighting conditions across views.
To create the training and test splits, we assign every 100th image in the dataset to the test set and use the rest for training.
Additionally, we test our method on the SeaThru-NeRF dataset~\cite{levy2023seathru}, a less challenging benchmark with only $\sim$25 front-view images per scene.

\noindent \paragraph{Implementation details.}
We implement our method based on 3DGS~\cite{kerbl20233d} for its simplicity and speed. 
We build a unified interface that integrates four UIR models~\cite{huang2025bayesian,fu2022uncertainty,han2022underwater,huang2023contrastive} to produce multiple sets of restored images.
The autoencoder is implemented with the code by~\citet{taesd2023}, but we change the latent dimension to $24\times32 \times 32$.
We directly borrow the masked decoder from~\cite{van2016conditional} to implement our illumination generator. 
Following 3DGS, the reconstruction loss $\mathcal{L}^{\text{rec}}$ consists of a combination of DSSIM and $L_1$ losses. Additionally, we jointly train the autoencoder and illumination generator in an end-to-end fashion. The final objective is given by
$
    \mathcal{L} = \mathcal{L}^\text{rec} +  \mathcal{L}^\text{contra} + \lambda_\text{ucn} \mathcal{L}^\text{ucn},
$
where $\lambda_\text{ucn}$ is set to 0.0005 in our experiments.
We disable the periodic opacity resetting (POR) originally used in 3DGS, as our uncertainty-aware opacity optimization renders it unnecessary.
We train our model for 15,000 and 30,000 iterations on the SeaThru-NeRF and BlueCoral3D datasets, respectively.

\subsection{Evaluation}

\paragraph{Baselines.} We compare our method against SeaThru-NeRF~\cite{levy2023seathru}, 3DGS~\cite{kerbl20233d}, WildGaussian~\cite{kulhanek2024wildgaussians}, UW-GS~\cite{wang2025uw}, GS-W~\cite{zhang2024gaussian}, and WaterSplatting~\cite{li2025watersplatting}.
As WildGaussian and GS-W are originally designed to handle appearance variations in uncontrolled images, they are potentially applicable to underwater illumination inconsistencies, and are therefore included as baselines in our evaluation.
To ensure reproducibility and fairness, all baseline results are obtained using official implementations.

\noindent \paragraph{Quantitative comparison.} 
The results on the Seathru-NeRF and BlueCoral3D datasets are provided in~\cref{tab:seathru_compare} and \cref{tab:blue3d_compare}, respectively. 
Models that do not account for dynamic illumination consistently yield lower scores. While WildGaussian and GS-W mitigate artifacts using 2D uncertainty masks, these strategies can be overly aggressive—often suppressing valid structures and harming reconstruction quality. In contrast, our R-Splatting models opacity uncertainty directly in 3D space via UAOO, resulting in significantly improved performance across all metrics.

Notably, our R-Splatting achieves a significantly larger performance gain on the BlueCoral3D dataset compared to the SeaThru-NeRF dataset. This is because BlueCoral3D features more pronounced illumination changes and complex camera motion, making it a more challenging setting—one for which R-Splatting is specifically designed.

\noindent \paragraph{Render Speed.} In Tab.~\ref{tab:seathru_compare}, we also compare rendering speed and training time. Since all views in our approach share a single $\mathbf{v}_i$ per point, the neural field only needs to be executed once for multi-view rendering, allowing R-Splatting to match 3DGS in rendering speed. In contrast, other methods with view-dependent neural fields require repeated evaluations, leading to slower rendering.

\begin{figure*}[h]
    \centering
    \includegraphics[width=0.95\linewidth]{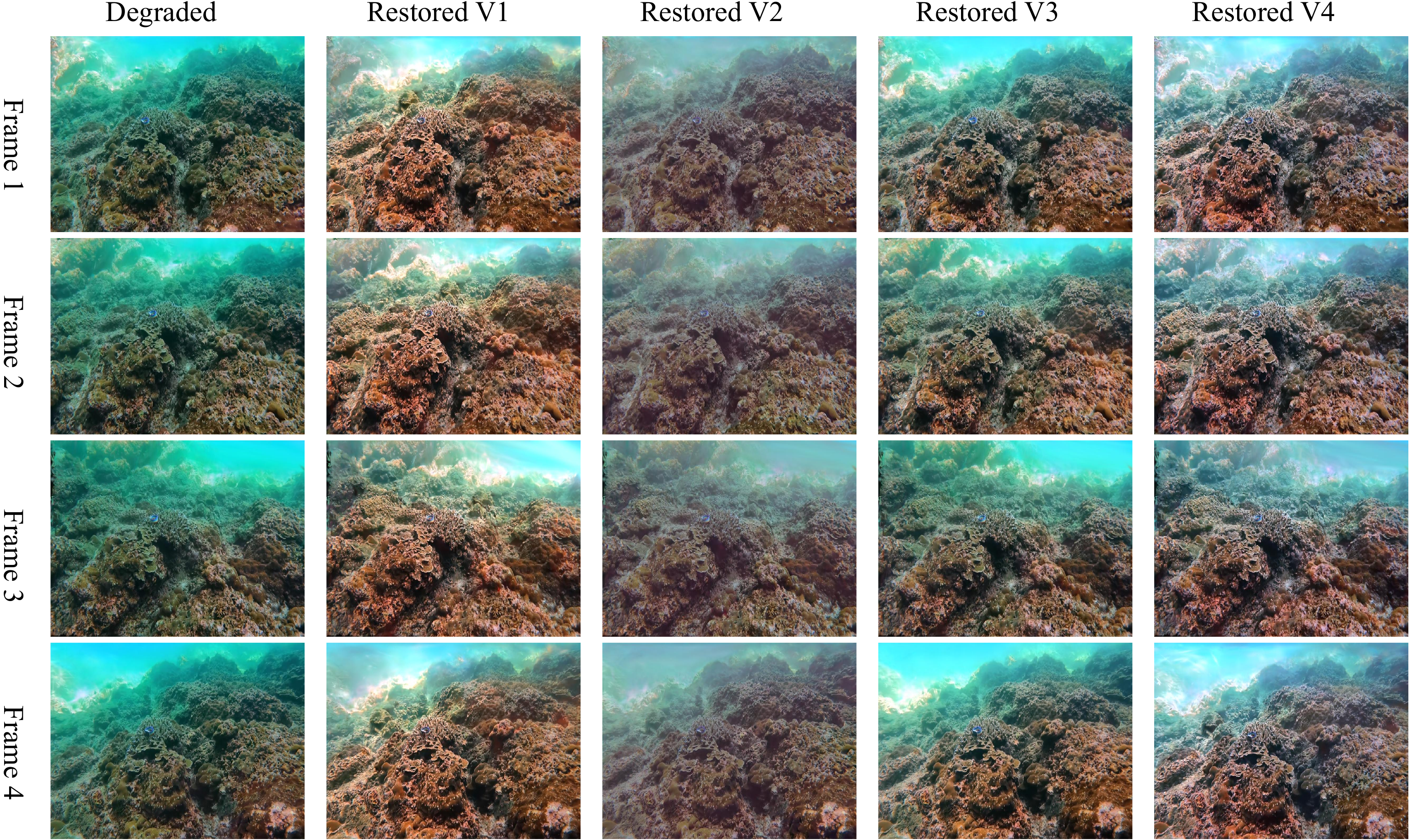}
    \caption{ \textbf{View-consistent renderings under diverse restoration styles.}
R-Splatting generates multiple restored outputs (V1–V4) for the same scene by sampling different latent codes from the illumination generator. While each column exhibits a unique restoration style, the renderings remain consistent across frames.}
    \label{fig:many}
\end{figure*}

\noindent\paragraph{Qualitative comparison} 
In~\cref{fig:visual_seathru_nerf}, we present a visual comparison between our method and existing baselines. 3DGS and GS-W fail to reconstruct certain structures and textures, primarily due to aggressive opacity resets and reliance on 2D mask-based filtering. Other methods also exhibit reduced visibility and missing content. In contrast, our R-Splatting consistently produces clearer results, successfully recovering even distant objects. Notably, in the second row, our method reconstructs the distant color calibration chart with higher fidelity than all baselines.

In~\cref{fig:visual_bluecoral3d}, we show that in shallow water scenes, refracted sunlight through surface waves produces complex illumination patterns (caustics). Methods like UW-GS and WaterSplatting, which rely on simplified physical models, fail to recover these effects. In contrast, R-Splatting benefits from the strong restoration capacity of UIR models and achieves significantly better rendering quality.

\noindent \paragraph{View-consistent renderings.} As shown in~\cref{fig:many}, our R-Splatting generates diverse restoration styles by sampling latent codes from the illumination generator. Owing to the view-invariant latent learning strategy (\cref{sec:view-invariant}), each style maintains both geometric and illumination consistency across views.

\subsection{Ablation Study}

We evaluate the contributions of individual components in R-Splatting by constructing a series of model variants, each selectively enabling or disabling Neural Field (NF), Uncertainty-Aware Opacity Optimization (UAOO), and Periodic Opacity Resetting (POR). Results are summarized in~\cref{tab:ablation}.

\begin{table}[h]
    \centering
    \small
    \setlength{\tabcolsep}{3pt}
    \begin{tabular}{l|ccc | ccc}
    Variant & NF  & UAOO  & POR  & \multicolumn{3}{c}{\textbf{Metrics}}  \\
     &   &  &  & PSNR & SSIM & LPIPS \\
    \hline
    M1 ( 3DGS) &  \xmark  & \xmark & \cmark & 22.81 & \cellcolor{lightyellow}0.810 & \cellcolor{lightyellow}0.278 \\
    M2 &  \cmark & \xmark & \cmark & \cellcolor{lightyellow}23.05  & \cellcolor{lightyellow}0.813  &  \cellcolor{lightred}0.265  \\
    M3 & \cmark  & \cmark & \cmark & 22.93  &  0.805 & 0.291 \\
    M4 & \xmark  & \cmark & \cmark & 22.71  &  0.798 & 0.306 \\
    M5 & \xmark  & \cmark & \xmark & \cellcolor{lightorange}24.32  &  \cellcolor{lightorange}0.817 & 0.282 \\
    M6 (R-Splatting)&  \cmark & \cmark & \xmark & \cellcolor{lightred}24.67  & \cellcolor{lightred}0.820  & \cellcolor{lightorange}0.277  \\
    \end{tabular}%
    \caption{The performance of different model variants on BlueCoral3D. We report the average scores of PSNR, SSIM and LPIPs of the dataset.}
    \label{tab:ablation}
\end{table}

Starting from the baseline 3DGS (M1), we first incorporate NF (M2), which improves PSNR from 22.81 to 23.05, suggesting the benefit of view-shared color modeling. In M3, we further introduce UAOO on top of M2, but observe a slight performance drop. To isolate this effect, M4 removes NF but retains UAOO and POR, resulting in a further performance degradation. When POR is disabled (M5), the performance improves significantly (PSNR 24.32), indicating that POR may conflict with the stochastic nature of UAOO. Finally, M6 combines NF and the UAOO–no-POR setting, achieving the best overall performance.

\section{Conclusion}
We present \textbf{R-Splatting}, a unified 3D reconstruction framework tailored for underwater scenes with diverse restoration styles and dynamic illumination. Our method introduces two key components: (1) \textit{Uncertainty-Aware Opacity Optimization} (UAOO) to robustly handle view-inconsistent lighting through stochastic opacity modeling, and (2) a \textit{conditional neural field} to enable view-shared color reconstruction from multiple UIR-enhanced inputs.
Compared to prior methods relying on simplified physical models, R-Splatting achieves superior geometric consistency and visual quality. Results on Seathru-NeRF and our newly collected \textbf{BlueCoral3D} dataset validate its effectiveness.

\section*{Acknowledgements}

This work was supported by the UKRI MyWorld Strength
in Places Program (SIPF00006/1) and the EPSRC ECR
International Collaboration Grants (EP/Y002490/1).

\bibliography{main}

\end{document}